\documentclass{l4dc2027}

\title[Conservatism of Latent Safety Filters]{Measuring the Conservatism of Latent Safety Filters for World Models}
\usepackage{times}
\usepackage{booktabs}
\usepackage{placeins}

\author{%
 \Name{Johannes Mootz} \Email{jmootz@ucsd.edu}\\
 \Name{Mani Amani} \Email{mamani@ucsd.edu}\\
 \addr University of California San Diego
 \AND
 \Name{Reza Akhavian} \Email{rakhavian@sdsu.edu}\\
 \addr San Diego State University%
}

\begin{document}

\maketitle

\begin{abstract}%
A latent safety filter decides from the latent of a world model whether a planned action is safe. 
Conformal calibration sets the threshold of the filter so that it accepts an unsafe action only with a small probability.
However, the filter can still be conservative and refuse many safe actions, depending on the latent.
To measure the conservatism that a latent forces on the filter, we group similar latents and actions into \emph{cells}.
A filter that gives one decision per cell can refuse the unsafe transitions of a cell only by refusing the safe ones in it as well.
We call the share of safe transitions that lie in such cells the \textit{overlap}.
We prove that the overlap is the minimum share of safe transitions that such a filter has to refuse in order to accept no unsafe transition in the data.
Both the overlap and the conformal calibration can be applied to any latent.
We use this to compare three existing designs, in which the safety labels train a classifier on top of the world model, the world model itself, or a separate encoder beside it.
In a visual navigation task, the separate encoder has the lowest overlap, and at the same conformal level its filter refuses the fewest safe actions and lets the planner reach the most goals of the three.
\end{abstract}


\section{Introduction}

World models let a robot plan from images and proprioception in a learned latent \citep{ha2018world,hafner2023mastering,toso2026preserving}.
However, they are prone to hallucinations \citep{hansen2026hallucination}, and a robot that relies on their predictions outside the training data can execute unsafe actions \citep{seo2025uncertainty,cao2026world}.
To improve safety, safety value functions from Hamilton--Jacobi reachability \citep{nakamura2025generalizing} and control barrier functions \citep{nakamura2025train} have been learned in the latent.
These latent safety filters check each proposed action and either switch to a backup policy \citep{nakamura2025generalizing,seo2025uncertainty,kim2025latent} or correct the action smoothly \citep{nakamura2025train}.

A latent safety filter is itself learned from data, so it is not fully reliable and can accept an action that leads to a failure \citep{seo2025uncertainty}.
Conformal calibration sets its threshold so that, under exchangeability, the probability of accepting a failure stays below a chosen level \citep{kim2025latent}.
This bound holds on any representation, and how often the filter refuses a safe action depends on the representation \citep{kim2025latent,feldman2026learning}.

A filter on a latent can tell apart only what the encoder keeps.
The encoder of a world model is trained to predict the next latent, and this objective can be minimized while the encoder discards every unstable mode of the system \citep{toso2026preserving} or encodes only features that are irrelevant for control \citep{toso2026learning}.
\citet{kim2025latent} show that a latent safety filter fails when the latent does not hold the safety-relevant information.
An encoder can map many states to the same latent, so a safe and an unsafe state can share one latent, and the filter then gives both the same decision.
A filter that is calibrated to rarely accept a failure refuses the safe one as well, which makes it conservative.

Reasoning about safety in a latent space is difficult, because a certificate on the latent does not automatically certify the true system.
\citet{lutkus2025latent} show that a latent barrier function keeps the true system safe if it keeps a margin to the unsafe set that is proportional to the error of the latent dynamics.
The set of states that an encoder maps to one latent is the \textit{fiber} of that latent, which is the nested pre-image of the latent under the observation map and the encoder.
\citet{mootz2026latent} show that the certificate needs a second margin, the fiber defect, which is how far safety can fall from one state of a fiber to another.
Both margins are computed from the true state, and in practice there are only logged transitions, each labeled safe or unsafe by its outcome.

In this work, we measure the conservatism that a latent forces on a safety filter from these labeled transitions.
On a continuous latent two logged transitions rarely share a latent, so we group similar latents and actions into \textit{cells}.
A filter that gives one decision per cell treats all transitions of a cell alike.
If a cell holds both safe and unsafe transitions, the filter can refuse the unsafe ones only by refusing the whole cell, so the safe ones are refused with them.
We call the share of safe transitions that lie in such cells the \textit{overlap}, and we prove that such a filter has to refuse at least this share to refuse every logged unsafe transition.
Figure~\ref{fig:lead} shows the cells of the latents of three encoders.

\begin{figure}[t]
\floatconts
  {fig:lead}
  {\caption{Cells of three latents for a ball that is pushed past a block, drawn over the position and speed of the ball. In the ground truth on the left, a push is safe in the green region and unsafe in the red one, and the block and the walls are gray. The top row shows the cells of each latent, and the bottom row colors a cell red if it holds an unsafe training transition and green otherwise. A cell that lies across the black boundary holds safe and unsafe states, so refusing it also refuses the safe ones.}}
  {\includegraphics[width=\linewidth]{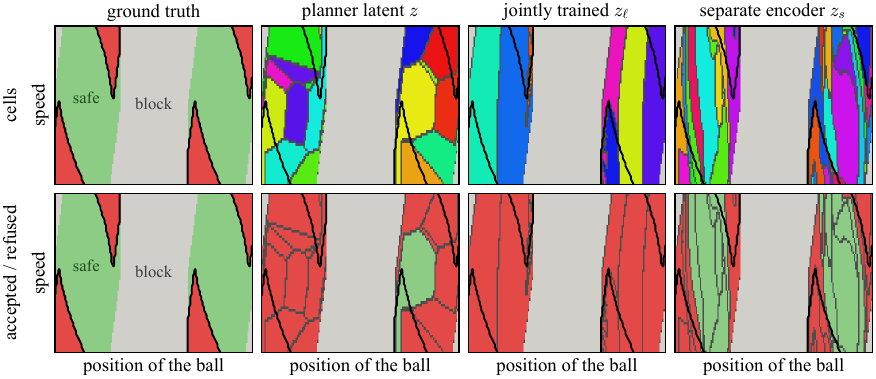}}
\end{figure}

The overlap and the conformal bound can be computed for any latent.
We use this to compare three existing designs of a latent safety filter, which differ in where the safety labels enter.
A classifier \citep{kim2025latent, nath2026pixels} or a projector \citep{agrawal2026anysafe} can be trained on top of the latent of the frozen world model.
The safety labels can be added to the training loss of the world model, so that the planner and the filter share one latent that was also trained to predict safety \citep{huang2024safedreamer,seo2025uncertainty}.
A separate encoder can be trained on the safety labels alone and placed beside the world model, which again stays unchanged \citep{tabbara2026designing,mani2025safety}.
We build all three for a visual navigation task with a world model and compare them at the same level of the conformal bound by their overlap and their refusals.
Our contributions are:
\begin{itemize}
\setlength{\itemsep}{2pt}
\setlength{\parskip}{0pt}
\setlength{\parsep}{0pt}
    \item We define the overlap of a latent using labeled transitions grouped into cells of similar latents and actions. It measures the share of safe transitions whose cell also contains an unsafe transition.
    \item We prove that the overlap is the minimum share of safe transitions that a filter with one decision per cell has to refuse in order to accept no unsafe transition in the data.
    \item We compare three existing designs in visual navigation with a fixed planner and the same conformal level. The separate safety encoder has the lowest overlap, and its filter refuses the fewest safe actions and reaches the most goals.
\end{itemize}

\section{Background}
\label{sec:background}

We consider a discrete-time dynamical system with state $x_t\in\mathcal{X}\subseteq\mathbb{R}^{n_x}$, control input $u_t\in\mathcal{U}\subseteq\mathbb{R}^{n_u}$ and deterministic dynamics
\begin{equation}\label{eq:dyn}
    x_{t+1} = f(x_t, u_t).
\end{equation}
The observation $y_t=g(x_t)\in\mathcal{Y}$ may combine images, proprioceptive and other measurements, and for encoders that use observation histories $x_t$ denotes the corresponding augmented state.
A world model provides an encoder $E:\mathcal{Y}\to\mathcal{Z}$ that maps an observation to its latent $z_t=E(y_t)$. We call the representation that an encoder produces its latent, and one point $z$ of it a latent value.
The controller observes $y_t$ and not $x_t$, and the planner operates on $z_t$.

\subsection{Safety Filters with a Backup Policy}
\label{sec:filters}

A safety filter checks a proposed action and switches to a backup policy when needed \citep{hsu2023safety}. 
Let $\mathcal{F}\subset\mathcal{X}$ be the failure set and $\pi_{\mathrm{b}}$ a stationary backup policy executable from the observations \citep{gurriet2020scalable}.
When one action spans several simulator steps, the state it reaches records whether its motion reached the failure set, so a violation during the action places $f(x,u)$ in $\mathcal{F}$.
Let $\mathcal{T}\subset\mathcal{X}$ be failure-free and invariant under $\pi_{\mathrm{b}}$ \citep{gurriet2020scalable}. 
From $x_0=x$ the backup rollout over a horizon of $H$ steps is $x_{i+1}=f(x_i,\pi_{\mathrm{b}}(x_i))$, and the recoverable set is
\begin{equation}\label{eq:C}
\mathcal{C}=\{x\in\mathcal{X}: x_0,\dots,x_H\notin\mathcal{F},\ x_H\in\mathcal{T}\}.
\end{equation}
This is the backup-set construction of \citet{gurriet2020scalable}.
The state-based filter accepts $u$ when $f(x,u)\in\mathcal{C}$ and applies $\pi_{\mathrm{b}}$ otherwise.

\subsection{Fibers and the Fiber Defect}
\label{sec:fibers}

Let $\Phi=E(g(x))$ map a state to its latent.
\citet{lutkus2025latent} show that a barrier function on the latent keeps the true system safe if it keeps a margin proportional to the error of the latent dynamics, which is the largest gap over a domain of states between the encoded true dynamics and the latent model.
For a safe set $\mathcal{S}_x\subseteq\mathcal{X}$, they define the induced latent safe set $\mathcal{S}_z=(\Phi(\mathcal{S}_x^{c}))^{c}$, the latent values to which no unsafe state is mapped. 
The fiber of a latent value $z$ is the set of states $x$ that are mapped to it,
\begin{equation}
\label{eq:fiber}
    \Phi^{-1}(z)=\{x\in\mathcal{X}:\Phi(x)=z\}.
\end{equation} 
A latent value is in $\mathcal{S}_z$ only if every state on its fiber is safe, so one unsafe state removes the whole fiber, and their latent barrier function certifies a subset of $\mathcal{S}_z$.
The states on one fiber can differ in their safety, and a filter on the latent cannot distinguish them \citep{mootz2026latent}.
For a safety function $h$, a reachable set $\mathcal{R}$ and a reference state $x_z$ on each fiber, \citet{mootz2026latent} define the fiber defect as
\begin{equation}
D(z)=\sup_{x\in\Phi^{-1}(z)\cap\mathcal{R}}[h(x_z)-h(x)].
\end{equation}
Every reachable state on the fiber then satisfies $h(x)\ge h(x_z)-D(z)$, so a certificate for the reference state carries over to the true state only with the margin $D(z)$. 
The defect needs the true state and the safety function, and Section~\ref{sec:defect} replaces it by a quantity that is counted from labeled transitions.

\subsection{Conformal Calibration}
\label{sec:conformal-calibration}

A learned filter rates every proposed action with a score and refuses it when the score is above a threshold.
Conformal calibration sets the threshold from the scores of held-out unsafe actions, so that a new unsafe action is accepted only with a chosen small probability $\delta$ \citep{vovk2005algorithmic,kim2025latent}.
Let $s_1,\dots,s_n$ be calibration scores and $s_{n+1}$ the score of a new point.
For $\delta\in(0,1)$ let $k=\lfloor\delta(n+1)\rfloor\ge1$ and let $\tau$ be the $k$-th smallest calibration score.

\begin{lemma}[{\citealp{vovk2005algorithmic}; \citealp[Lemma~1]{lindemann2025formal}}]\label{lem:conformal}
If $s_1,\dots,s_{n+1}$ are exchangeable, then $\mathbb{P}(s_{n+1}<\tau)\le\delta$, over the draw of all $n+1$ scores.
\end{lemma}
The scores are exchangeable when the calibration points and the new point are drawn independently from one distribution and the score was fixed before the calibration points were seen.
The bound is marginal over the calibration data and the new point, not a guarantee for each fixed calibration set \citep{lindemann2025formal}.
Conformal calibration is also used for warning systems on learned image features \citep{feldman2026learning}, and \citet{huriot2026safe} replace it by adaptive conformal inference, which bounds a long-run average.

\section{Problem Formulation}

Consider the system~\eqref{eq:dyn}, where a fixed world-model planner proposes an action $\hat u_t=\pi_{\mathrm{p}}(E(y_t))$.
We seek a safety filter $r$ that selects
\begin{equation}
u_t=
\begin{cases}
\hat u_t, & r(y_t,\hat u_t)=0,\\
\pi_{\mathrm{b}}(y_t), & r(y_t,\hat u_t)=1.
\end{cases}
\end{equation}
For proposals from starts in $\mathcal{C}$, the filter should keep the probability of accepting a proposal that leaves $\mathcal{C}$ below a chosen level $\delta$ and refuse as few proposals that stay in $\mathcal{C}$ as possible.
The filter's input is $y_t$ rather than $x_t$, so what it can decide depends on the representation it uses.
Given labeled outcomes of proposed actions followed by the backup, we ask how much the representation forces the filter to refuse, and how this differs between the ways of encoding safety in the representation.

\section{Safety Filters on a Latent}
\label{sec:method}

A latent filter uses the observation only through the latent $z_t=E(y_t)$, so it is a function of the latent and the proposed action,
\begin{equation}\label{eq:readout}
r:\mathcal{Z}\times\mathcal{U}\to\{0,1\},
\qquad r(z,u)=0 \text{ accepts }u,
\qquad r(z,u)=1 \text{ applies }\pi_{\mathrm{b}}.
\end{equation}
The outcome label records whether the backup recovers after the proposed action,
\begin{equation}\label{eq:labels}
\ell(x)=\mathbf{1}[x\notin\mathcal{C}], \qquad
\ell^{+}(x,u)=\ell\big(f(x,u)\big),
\end{equation}
and a transition with $\ell^{+}=1$ is a hazard.
The labels are computed offline from a rollout, and the filter uses no state estimate at run time.

\subsection{Contaminated Fibers} 
\label{sec:readouts}

A latent filter sees the state only through the latent, so two states with the same latent receive the same decision for every action.
\begin{definition}\label{def:mixed}
A fiber $\Phi^{-1}(z)$ is \emph{contaminated for} $u$ if it contains $x$ and $x'$ with $\ell^{+}(x,u)\neq\ell^{+}(x',u)$.
\end{definition}

\begin{proposition}\label{prop:readout}
Let $r$ be any function of $(z,u)$.
\begin{enumerate}
\item[(a)] If $\Phi(x)=\Phi(x')$ then $r(\Phi(x),u)=r(\Phi(x'),u)$ for every $u$.
On a fiber contaminated for $u$, every such $r$ is wrong on at least one member.
\item[(b)] If $g(x)=g(x')$ then $\Phi(x)=\Phi(x')$. Every fiber of $\Phi$ is a union of fibers of $g$.
\end{enumerate}
\end{proposition}
\begin{proof}
(a) $r$ depends on $x$ only through $\Phi(x)$. (b) $\Phi=E\circ g$.
\end{proof}

Part (a) is the statement of Corollary 1 of \citet{mootz2026latent} applied to a binary outcome after one action.
A safety value on the predicted latent, a latent barrier function and a monitor of the model error are functions of $(z,u)$ as well \citep{nakamura2025generalizing,nakamura2025train,lutkus2025latent,kim2025latent}, so no filter downstream of the encoder can tell which member of a contaminated fiber it is in.
Part (b) separates the information absent from the observation from the information the encoder drops, and a second encoder can only recover the latter.
For a continuous latent we fix a cell assignment map $q:\mathcal{Z}\to\{1,\dots,K_z\}$ and use the coarsened encoder $\bar\Phi=q\circ\Phi$.
Its fibers are the preimages of the latent cells, so Proposition~\ref{prop:readout} holds for $\bar\Phi$ as well.

\subsection{The Overlap of a Latent}
\label{sec:defect}

How often a filter is wrong on a contaminated fiber depends on the data, and we count it on a finite set $\mathcal{D}$ of labeled training transitions such that $
\mathcal{D}
=
\left\{
\left(x_i,u_i,\ell_i^{+}\right)
\right\}_{i=1}^{N}.
$
A fiber is contaminated when two transitions have the same latent and the same action and different labels.
A neural encoder on images gives every logged transition its own value of $(z,u)$, because the latent and the action are continuous.
No two logged transitions then lie on the same fiber, so the data cannot show a contaminated fiber, even when transitions that are close differ in their label.
The count therefore needs a resolution at which transitions that are close count as the same.
We fix it by a partition of $\mathcal{Z}\times\mathcal{U}$, the cells of $\bar\Phi$ crossed with $K_u$ bins of the action.
The action is binned too, because the same state can be safe under one action and unsafe under another, so only transitions with similar actions should be compared.
The bins only decide which logged transitions are compared, and they do not restrict the actions that a filter can accept.
With a map $b:\mathcal{U}\to\{1,\dots,K_u\}$ that assigns each action to a bin, a \emph{cell} is the set $\mathcal{Q}_{ik}=\{(z,u)\in\mathcal{Z}\times\mathcal{U}: q(z)=i,\ b(u)=k\}$, the pairs in latent cell $i$ and action bin $k$.

A filter at this resolution gives one decision per cell, so every latent and action in the cell is treated the same.
A filter with a trained classifier does not give one decision per cell, because the classifier reads the latent itself and can decide differently within a cell.
We compare latents at the same resolution.
On $\mathcal{D}$, $\operatorname{miss}(r)$ is the share of hazards accepted and $\operatorname{refusals}(r)$ the share of safe transitions refused.

\begin{definition}\label{def:overlap}
For fixed data $\mathcal{D}$ and a fixed partition of $\mathcal{Z}\times\mathcal{U}$, the \emph{overlap} $D_0(\Phi)$ is the share of safe transitions whose cell also holds a hazard.
\end{definition}

\begin{proposition}\label{prop:overlap}
For filters $r$ that give one decision per cell of a fixed partition of $\mathcal{Z}\times\mathcal{U}$, counted on $\mathcal{D}$,
\begin{enumerate}
\item[(a)] every $r$ with $\operatorname{miss}(r)=0$ has $\operatorname{refusals}(r)\ge D_0(\Phi)$,
\item[(b)] the filter that refuses exactly the cells holding a hazard has $\operatorname{miss}=0$ and $\operatorname{refusals}=D_0(\Phi)$.
\end{enumerate}
\end{proposition}
\begin{proof}
Group the transitions of $\mathcal{D}$ by cell, cell $j$ holding $a_j$ safe transitions and $b_j$ hazards, so that $r$ gives one verdict per cell.
(a) $\operatorname{miss}(r)=0$ forces every cell with $b_j\ge1$ to be refused, so the number of refused safe transitions is at least $\sum_{b_j\ge1}a_j$, and dividing by $\sum_j a_j$ gives $D_0(\Phi)$.
(b) Refusing exactly these cells accepts no hazard and refuses exactly these $\sum_{b_j\ge1}a_j$ safe transitions.
\end{proof}
The overlap depends on the encoder, the partition and the data.
With the label as safety function, the fiber defect of \citet{mootz2026latent} is one as soon as a single fiber is contaminated, so it cannot rank latents.

\subsection{A Conformal Certificate on any Latent}
\label{sec:conformal-certificate}

On fresh transitions we allow the filter a miss rate $\delta$ and use a score $s(z,u)$ with the filter $r_\tau(z,u)=\mathbf{1}[s(z,u)\ge\tau]$.
The threshold is the conformal threshold at level $\delta$ on $n$ calibration hazards, transitions with $\ell^{+}=1$ from starts in $\mathcal{C}$, held out from $\mathcal{D}$ and from any data the score or the encoder was trained on.
A refusal is safe only if the backup keeps the system in $\mathcal{C}$.

\begin{lemma}\label{lem:invariant}
Let $f$ be deterministic, $\pi_{\mathrm{b}}$ a stationary function of the state, and $\mathcal{T}$ failure-free and invariant under $\pi_{\mathrm{b}}$.
Then $x\in\mathcal{C}$ implies $f(x,\pi_{\mathrm{b}}(x))\in\mathcal{C}$.
\end{lemma}
\begin{proof}
For $x\in\mathcal{C}$ the backup rollout $x_1,\dots,x_H$ avoids $\mathcal{F}$ and ends in $\mathcal{T}$, and $x_{H+1}=f(x_H,\pi_{\mathrm{b}}(x_H))$ lies in $\mathcal{T}$ by invariance, so it is failure-free.
The rollout from $x_1=f(x,\pi_{\mathrm{b}}(x))$ is the same sequence without its first element, hence it avoids $\mathcal{F}$ for $H$ steps and ends in $\mathcal{T}$.
\end{proof}
This is the invariance of the backup-set construction \citep{gurriet2020scalable,cosner2021measurement}.
With the threshold set on the held-out hazards, Lemma~\ref{lem:conformal} bounds the misses of the score on the latent, and with Lemma~\ref{lem:invariant} this becomes a bound on leaving $\mathcal{C}$.

\begin{theorem}\label{thm:cert}
Consider proposed transitions starting in $\mathcal{C}$.
Let the score $s$ and the encoder $E$ be fixed before the $n$ calibration hazards are seen.
Conditional on the new transition being a hazard, let its score and the $n$ calibration scores be jointly exchangeable.
Let $\delta\in(0,1)$ with $n\ge1/\delta-1$, and let $\tau$ be the conformal threshold at level $\delta$.
Then, over the calibration draw and the new transition,
\begin{enumerate}
\item[(i)] $\mathbb{P}(r_\tau=0\mid\ell^{+}=1)\le\delta$,
\item[(ii)] under the hypotheses of Lemma~\ref{lem:invariant}, accepting $u$ when $r_\tau=0$ and applying $\pi_{\mathrm{b}}$ otherwise leaves $\mathcal{C}$ in one step with probability at most $\delta$.
\end{enumerate}
\end{theorem}
\begin{proof}
(i) is Lemma~\ref{lem:conformal} applied to the hazard scores similar to \citet{kim2025latent}.
For (ii), a refused action keeps the state in $\mathcal{C}$ by Lemma~\ref{lem:invariant}, so leaving $\mathcal{C}$ requires accepting a hazard, which has probability at most $\delta$ by (i).
\end{proof}

\section{Learning a Latent Safety Filter}
\label{sec:learning}

\subsection{Three Designs of a Latent Safety Filter}
\label{sec:separate}

The certificate of Section~\ref{sec:method} holds for every encoder at the same level $\delta$, but the overlap differs from one encoder to the next.
The overlap is low when the encoder keeps safe transitions and hazards apart, and this depends on the encoder.
Existing latent safety filters differ in where the safety labels enter the training, and we evaluate three designs.
In the first design the world model is trained without safety labels and stays unchanged, and a classifier $s(z,u)$ is trained with the labels on top of its latent $z=E(y)$, the planner latent \citep{kim2025latent}, or a projector of this latent \citep{agrawal2026anysafe}.
The classifier sees only the latent and the action, so a safe transition and a hazard that agree in both get the same score (Proposition~\ref{prop:readout}).
A better classifier can separate transitions within a cell, but it cannot repair a contaminated fiber.

In the second design the safety labels enter the training of the world model itself \citep{huang2024safedreamer,seo2025uncertainty}.
A predictor of the label from the latent is added to the world model, and its loss is added to the training loss of the world model.
The encoder is thereby pushed to keep what this predictor needs, and the planner and the filter share the resulting latent $z_\ell$.
In the third design the world model again stays unchanged, and a second encoder $E_s$ is trained on the safety labels alone \citep{tabbara2026designing,mani2025safety}.
The safety encoder maps the observation to a safety latent $z_s=E_s(y)$.
A classifier $s(z_s,u)$ predicts the label from this latent and the action.
The planner proposes an action from $z=E(y)$ and the filter checks it from $z_s$, so the planner and its dynamics model are not affected.
The safety encoder reads the observation, so it can keep what $E$ drops, but not what the observation lacks (Proposition~\ref{prop:readout}).

\subsection{Measuring and Certifying the Learned Latent}
\label{sec:diagnosing}

To compare the three designs, we count the overlap of each latent and build a calibrated filter on it.
The overlap of a latent needs only logged transitions with their labels $\ell^{+}$.
The filter needs in addition a score $s(z,u)$ on the latent, for example a classifier, a safety value or a barrier.
Its certificate holds when the encoder and the score are fixed before calibration and the calibration hazards are exchangeable with the hazards encountered in use.
These requirements are the same for every latent, so every latent gets the same bound on its misses, and what differs is how many safe actions its filter refuses.
We divide the latent and the action into cells and count the overlap $D_0$.
A filter that gives one decision per cell must refuse at least this share of the safe transitions if it accepts no hazard.
However, the overlap only describes the latent, and this filter is too coarse to deploy, since a single hazard in a cell makes it refuse all transitions of that cell.
The filter we deploy therefore does not use the cells.
It uses a trained classifier as its score, which reads the latent and the action themselves and can separate safe transitions from hazards within a cell.
By Theorem~\ref{thm:cert} the calibrated filter accepts a hazard with probability at most $\delta$, and this holds for any classifier, even a poorly trained one.
A worse classifier has the same bound on its misses, but it typically refuses more safe actions.

\section{Experiments}
\label{sec:application}

We now check how conservative the three designs are.
A robot knows its position and speed from its own sensors, but an obstacle in its way appears only in the camera image, and a filter on a latent can react to it only if the encoder keeps it.
We build this situation in the visual PointMaze of \citet{toso2026preserving} in MuJoCo, where a planner steers a ball through a U-shaped maze to a goal (Figure~\ref{fig:setting}).
The encoders receive two consecutive camera images and the position and velocity of the ball, which stand for the camera and the motion sensors of the robot.
We place one block at a random position in the maze, which is visible in the images and does not appear in the position and velocity of the ball.
The filter checks one action at a time, which is five consecutive pushes on the ball, and when it refuses the action the ball brakes instead.
An action is a hazard if the ball touches a wall or the block during the pushes or the brake that follows, and safe otherwise.
The data are 3000 episodes of random actions, split by episode into training, calibration and test data (Appendix~\ref{app:env}).

\begin{figure}[t]
\floatconts
  {fig:setting}
  {\caption{The setting, shown on one scene pair. The two scenes differ only in where the block sits. The camera image is what the encoders receive, at 64 by 64 pixels. In the top view the walls are gray, the block is orange, the dashed squares are the other positions a block can take, and the arrow is the velocity of the ball.}}
  {\includegraphics[width=\linewidth]{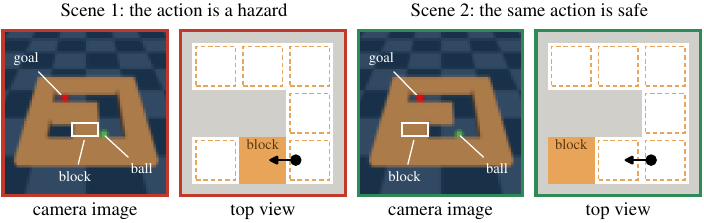}}
\end{figure}

The three designs are built on one world model (Appendix~\ref{app:models}).
For the first design, the planner latent $z$ is the latent of the world model of \citet{toso2026preserving}, which we train on our data without the safety labels.
For the second design, the jointly trained latent $z_\ell$ comes from the same model with a label predictor added to its training.
For the third design, the safety latent $z_s$ comes from a separate convolutional encoder with a latent of 20 numbers, trained on the labels alone.
The planner is the same for all three designs. 
It plans on the original world model and proposes an action, and the filter checks this action from its own latent before it is executed.
To turn each latent into a filter, we train the same small classifier on it and set its threshold by conformal calibration at the level $\delta=0.02$.
We first count the overlap of each latent and then measure the refusals of its filter on logged transitions.
We then test with scene pairs whether each latent keeps the block, and finally run the planner with each filter.
A scene pair is two scenes with the same ball, goal and action, in which only the block sits in a different place, so that the action is a hazard in one scene and safe in the other (Figure~\ref{fig:setting}).

\begin{table}[t]
\centering
\caption{
The three designs on the logged transitions and the 600 scene pairs.
All entries are shares between 0 and 1.
The overlap is a range over three partitions, and bold marks the best value. 
}
\label{tab:representations}
{\small\setlength{\tabcolsep}{3pt}
\begin{tabular}{lcccccc}
\toprule
& \multicolumn{2}{c}{Measured on cells}
& \multicolumn{4}{c}{Calibrated filter, $\delta=0.02$} \\
\cmidrule(lr){2-3}\cmidrule(lr){4-7}
Latent
& \shortstack{Overlap\\$D_0$ $\downarrow$}
& \shortstack{Scene pairs\\in one cell $\downarrow$}
& \shortstack{All hazards\\accepted}
& \shortstack{Block hazards\\accepted $\downarrow$}
& \shortstack{Scene pairs\\separated $\uparrow$}
& \shortstack{Safe actions\\refused $\downarrow$} \\
\midrule
Planner latent $z$
& 0.96--0.97 & 1.00 & 0.018 & 0.153 & 0.00 & 0.302 \\
Jointly trained latent $z_\ell$
& 0.94--0.96 & 0.98 & 0.021 & 0.116 & 0.10 & 0.318 \\
Safety latent $z_s$
& \textbf{0.44--0.46} & \textbf{0.09} & 0.021 & \textbf{0.029} & \textbf{0.86} & \textbf{0.161} \\
\bottomrule
\end{tabular}}
\end{table}

\paragraph{Overlap.}
We measure the overlap on the training transitions, with each latent divided into $K_z=100$ cells by k-means and the action into $K_u=9$ bins (Appendix~\ref{app:columns}).
On the planner latent and on the jointly trained latent almost every safe transition lies in a cell that also holds a hazard, while on the safety latent fewer than half do (Table~\ref{tab:representations}).
Figure~\ref{fig:lead} shows the cells for one action.
The cells of the first two latents lie across the true boundary between safe and hazard, and the cells of the safety latent follow it.
A filter with one decision per cell would therefore have to refuse almost everything on the first two latents (Proposition~\ref{prop:overlap}).

\paragraph{Calibrated filters.}
Each of the three calibrated filters accepts about $\delta$ of the hazards, in line with Theorem~\ref{thm:cert}, so the three are compared at the same rate of misses.
At this rate the filter on the safety latent refuses about half as many safe actions as the other two.
The design with the lowest overlap is therefore also the one whose trained filter is least conservative.
These numbers compare the classifiers we trained and do not establish the best possible classifier on each latent.
At the other levels the filter on the safety latent also refuses least, except at $\delta=0.2$ where the three are equal (Appendix~\ref{app:columns}).

\paragraph{Scene pairs.}
The scene pairs test whether a latent keeps the block, because the two scenes of a pair differ in nothing else.
The planner latent does not keep it, since both scenes of every pair fall into one cell and its filter separates none of them (Table~\ref{tab:representations}).
The block is in the images that this encoder receives, but its latent does not hold it.
The jointly trained latent keeps little more, although the labels were part of its training.
The safety latent keeps the block.
A filter that does not see the block still meets the bound on all hazards, but it accepts a much larger share of the hazards that the block causes.

\paragraph{Planner.}
We now run the planner with each filter to see whether fewer refusals let it reach more goals (Table~\ref{tab:solvable}).
An episode has a random start, goal and block position, the same for every filter, and it ends at the goal, at the first contact, or after 60 actions, when the filter has stopped the planner.
In this experiment the filter on the safety latent uses the classifier that was trained with its encoder (Appendix~\ref{app:planner}).
Without a filter the planner reaches most goals, but many episodes end in a contact.
The filters on the planner latent and the jointly trained latent refuse many safe actions, and this stops the planner before most goals.
The filter on the safety latent removes most contacts and reaches as many goals as the planner without a filter, as does the reference filter on the full state, which reads the position of the block from the simulator.
With the filter on the safety latent more episodes end in a wall contact than with the other filters.

\begin{table}[t]
\centering
\caption{
The unchanged planner with each filter at $\delta=0.02$.
Backup used, stopped and goals reached are shares of the episodes with a reachable goal, and contacts are shares of all episodes (Appendix~\ref{app:planner}).
Bold marks the best design.
}
\label{tab:solvable}
\begin{tabular}{lccccc}
\toprule
Filter
& \shortstack{Backup\\used $\downarrow$}
& \shortstack{Stopped $\downarrow$}
& \shortstack{Goals\\reached $\uparrow$}
& \shortstack{Wall\\contacts $\downarrow$}
& \shortstack{Block\\contacts $\downarrow$} \\
\midrule
None & 0.00 & 0.00 & 0.72 & 0.37 & 0.11 \\
Full state $x$ (reference)& 0.50 & 0.22 & 0.74 & 0.03 & 0.00 \\
\midrule
Planner latent $z$
& 0.79 & 0.61 & 0.39 & 0.01 & 0.03 \\
Jointly trained latent $z_\ell$
& 0.81 & 0.68 & 0.32 & 0.00 & 0.01 \\
Safety latent $z_s$
& \textbf{0.57} & \textbf{0.22} & \textbf{0.72} & 0.06 & 0.01 \\
\bottomrule
\end{tabular}
\end{table}

\section{Conclusion}
\label{sec:conclusion}

In this work, we measured how conservative a latent forces a safety filter to be, using only logged transitions that are labeled safe or unsafe.
We defined the overlap, the share of safe transitions that lie in a cell together with an unsafe one, and proved that no filter with one decision per cell can refuse less without accepting an unsafe transition of the data.
The overlap and the conformal bound apply to any latent, so we used them to compare three existing designs of a latent safety filter at the same level.
On a visual world model the separate safety encoder had the lowest overlap, and its filter refused fewer safe actions, stopped the planner less often and reached more goals, while adding the labels to the world model changed little.
These results are from one environment and one world model.
Future work will test the comparison on other world models and on hardware, and extend the conformal bound from the random actions of the logged data to the actions that a planner proposes.

\clearpage
\bibliography{references}

\subsection*{Generative AI Use}
Generative AI tools were used for language editing, including improving grammar and spelling. 
The authors reviewed and verified all AI-assisted content and take responsibility for the final content of the paper.

\clearpage
\appendix
\section{Experimental Details}
\label{app:details}

\FloatBarrier
\subsection{Environment, Data and Labels}
\label{app:env}

Table~\ref{tab:env} lists the environment, the data and the labels.

\begin{table}[!htb]
\floatconts{tab:env}{\caption{Environment, data and labels.}}{%
\begin{tabular}{p{0.16\linewidth}p{0.77\linewidth}}
\toprule
Environment & PointMaze U, $5\times5$ map with seven free cells. One extra wall cell, the block, is placed uniformly in one of them in each episode. \\
State & Ball position and velocity $x_{\mathrm{prop}}\in\mathbb{R}^4$. The full state $x$ adds the block position and the position of the ball relative to the block. \\
Observation & Current and previous $64\times64$ RGB frames and $x_{\mathrm{prop}}$. \\
Action & Five pushes in $[-1,1]^2$, one per simulator step. The classifier reads all ten components, and the action bins use their two-dimensional sum. \\
Data & 3000 episodes of 60 random actions. Episodes 0--1599 for training, 1600--1999 for development (choice of checkpoints and settings), 2000--2599 for calibration and 2600--2999 for testing. The last 1000 were generated after all models and settings were fixed. \\
Backup & Brake $u=\operatorname{clip}(-3v,-1,1)$ for $H=10$ actions. After these actions the ball had zero speed and no later contact in all 2500 rollouts we checked.\\
Labels & An action is a hazard if the ball touches a wall or the block during the five pushes or during the $H$ brake actions that follow. 16.8\,\% of the logged transitions are hazards. All encoders and classifiers were trained on the same earlier labels, which did not yet count a contact during the pushes, and all are evaluated on the labels above.\\
\bottomrule
\end{tabular}}
\end{table}

\FloatBarrier
\subsection{Encoders of the three Designs}
\label{app:models}

Table~\ref{tab:three-models} lists the encoder, the latent and the training loss of each design.
The planner latent and the jointly trained latent are one training run each, and the safety encoder was trained with three seeds.

\begin{table}[!htb]
\floatconts{tab:three-models}{\caption{The three designs.}}{%
\small
\begin{tabular}{p{0.11\linewidth}p{0.26\linewidth}p{0.26\linewidth}p{0.26\linewidth}}
\toprule
 & Planner latent $z$ & Jointly trained $z_\ell$ & Safety latent $z_s$ \\
\midrule
Input & Current frame, previous frame and their difference, and $x_{\mathrm{prop}}$. & Same as $z$. & Same as $z$. \\
Encoder & ViT, four blocks, width 192, patch size 8. The class token and $x_{\mathrm{prop}}$ pass through an MLP. & Same as $z$. & Four convolutional blocks (32, 64, 128, 128 channels) and a fully connected layer of width 256. \\
Latent & 192 numbers. & 192 numbers. & 20 numbers, 16 image features and $x_{\mathrm{prop}}$. \\
Predictor & Six-layer predictor of the next latent. & Predictor of $z$ and a label predictor on the latent and the action, two layers of width 256. & One layer of width 64 on the latent and the action, with a hazard output and a brake-time output. \\
Loss & Next-latent prediction and endpoint inverse dynamics. & Loss of $z$ plus cross-entropy on the safety label with weight one. & Cross-entropy on the safety label and a Huber loss on the brake time. \\
\bottomrule
\end{tabular}}
\end{table}

\FloatBarrier
\subsection{Details on the Scene Pairs}

A scene pair is one position and velocity of the ball and one action, simulated with the block in two different maze cells.
The positions come from a grid in the free cells near the block with offsets of $\pm0.3$, and the velocities range from $-2$ to $2$ on each axis.
We label every combination under both block positions and keep those whose start is in $\mathcal{C}$ under both and whose action is a hazard under one and safe under the other.
This gives 1794 candidates, from which we draw 600 pairs at random.
For each scene we render the previous frame, apply five simulator steps without a push and render the current frame, so the proprioceptive states are the same in the two scenes.
On standardized latents, the median distance between the two scenes of a pair is $1.2\cdot10^{-5}$ of the median distance between the hazard scenes of different pairs on $z$, 0.011 on $z_\ell$ and 0.88 on $z_s$.
The score of the classifier on $z$ differs by at most $1.6\cdot10^{-4}$ between the two scenes, so its filter gives both the same decision. It refuses both in 484 of the 600 pairs and accepts both in 116.

\FloatBarrier
\subsection{Details on Table~\ref{tab:representations}}
\label{app:columns}

Table~\ref{tab:columns} states how each column of Table~\ref{tab:representations} is computed, and Table~\ref{tab:heads-sweep} gives the safe actions refused at other levels.
The first two rows of Table~\ref{tab:heads-sweep} are references, the same classifier on the position and velocity of the ball alone and on the full state.
The rate of accepted hazards is computed per test episode and then averaged, which is the probability that Theorem~\ref{thm:cert} bounds for a hazard drawn as in the calibration.
The two rates above $\delta$ in Table~\ref{tab:representations} lie within two standard errors of $\delta$.
With $K_z=50$ the overlap is 0.97--1.00 on $z$, 1.00 on $z_\ell$ and 0.52--0.54 on $z_s$, and with $K_z=200$ it is 0.84--0.86, 0.78--0.81 and 0.37.
A filter with one decision per cell, fitted on the training transitions, refuses 0.42--0.44 of the safe test transitions on $z_s$ and accepts 0.011--0.013 of the test hazards.
Over the 200 calibration draws the 10th to 90th percentile of the safe actions refused at $\delta=0.02$ is 0.26--0.34 on $z$, 0.28--0.36 on $z_\ell$ and 0.14--0.18 on $z_s$.
With the classifier trained with its encoder, the filter on $z_s$ refuses 0.194 of the safe test actions and separates 0.79 of the scene pairs.

\begin{table}[!htb]
\floatconts{tab:columns}{\caption{How the columns of Table~\ref{tab:representations} are computed.}}{%
\small
\begin{tabular}{p{0.2\linewidth}p{0.73\linewidth}}
\toprule
Cells & K-means with $K_z=100$ clusters on 96,000 training transitions per representation, crossed with $K_u=9$ bins of the net push. Three partition seeds. \\
Overlap & Share of the safe training transitions whose cell holds one of the training hazards. \\
Classifier & MLP with hidden widths 128 and 64 on the standardized representation and the ten action components. \\
Threshold & One hazard with a start in $\mathcal{C}$ from each of the 598 calibration episodes that hold one, and $\tau$ the 11th smallest score ($\delta=0.02$). 200 draws. \\
Safe actions refused & Share of the safe test transitions with a score of at least $\tau$, mean over the draws. \\
Hazards accepted & Share of the test hazards with a start in $\mathcal{C}$ that score below $\tau$, per episode and then averaged. A block hazard is a hazard that would not be one without the block, and for block hazards the share is taken among them. \\
Scene pairs & Cells of the first partition seed and the threshold of the median draw. A pair is in one cell when both of its scenes fall into the same cell. A pair is separated when the filter refuses the action in the hazard scene and accepts it in the safe scene. \\
\bottomrule
\end{tabular}}
\end{table}

\begin{table}[!htb]
\floatconts{tab:heads-sweep}
{\caption{Safe actions refused on the test data across calibration levels, means over 200 calibration draws.}}
{%
\begin{tabular}{lccccccc}
\toprule
Representation & $\delta=0.002$ & $0.005$ & $0.01$ & $0.02$ & $0.05$ & $0.1$ & $0.2$ \\
\midrule
Proprioceptive state $x_{\mathrm{prop}}$ & 0.840 & 0.780 & 0.692 & 0.556 & 0.365 & 0.199 & 0.082 \\
Full state $x$ & 0.600 & 0.496 & 0.377 & 0.259 & 0.163 & 0.092 & 0.050 \\
\midrule
Planner latent $z$ & 0.668 & 0.555 & 0.402 & 0.302 & 0.195 & 0.105 & \textbf{0.041} \\
Jointly trained latent $z_\ell$ & 0.643 & 0.543 & 0.427 & 0.318 & 0.196 & 0.109 & 0.042 \\
Safety latent $z_s$ & \textbf{0.280} & \textbf{0.247} & \textbf{0.205} & \textbf{0.161} & \textbf{0.114} & \textbf{0.072} & 0.044 \\
\bottomrule
\end{tabular}}
\end{table}

\FloatBarrier
\subsection{Details on Table~\ref{tab:solvable}}
\label{app:planner}

Table~\ref{tab:planner-details} gives the settings of the planner experiment.

\begin{table}[!htb]
\floatconts{tab:planner-details}{\caption{The planner experiment.}}{%
\small
\begin{tabular}{p{0.2\linewidth}p{0.73\linewidth}}
\toprule
Planner & CEM on the unchanged world model with $z$: horizon 5, 300 samples, goal given as an encoded image. It executes its five-action plan and plans again when the plan is used up or after a brake. Each action is one push held for the five simulator steps. \\
Filter & Checks every proposed action, and on a refusal the brake is applied for the five simulator steps. On $x$, $z$ and $z_\ell$ it uses the classifier of Table~\ref{tab:columns}, on $z_s$ the classifier trained with its encoder. Threshold of the median calibration draw. \\
Episodes & Three runs of 50, one seed of the safety encoder per run. Block cell, start and goal are drawn at random from the episode seed, the same for every filter. An episode is counted up to the goal (within a distance of 0.5), the first contact or 60 actions. \\
Reachable goal & Start and goal lie on the same side of the block, in 109 of the 150 episodes. \\
Goals reached, stopped & Shares of the 109 episodes that reach the goal before any contact, or neither the goal nor a contact within 60 actions. \\
Backup used & Share of the 109 episodes in which the filter braked at least once. \\
Contacts & Shares of all 150 episodes that end in a contact with a wall or with the block. \\
\bottomrule
\end{tabular}}
\end{table}

\end{document}